\documentclass[10pt,journal,compsoc]{IEEEtran}
\usepackage{subfigure}

\ifCLASSOPTIONcompsoc
  \usepackage[nocompress]{cite}
\else
  \usepackage{cite}
\fi
\ifCLASSINFOpdf
  \usepackage[pdftex]{graphicx}
  \graphicspath{{images/}}
\else
\fi
\usepackage{color}
\usepackage{multirow}

\usepackage{array}
\newcolumntype{C}[1]{>{\centering\let\newline\\\arraybackslash\hspace{0pt}}m{#1}}

\usepackage{times}
\usepackage{textcomp}
\usepackage{makecell}
\usepackage{url}

\usepackage{amsmath}
\usepackage{commath}
\usepackage{dsfont}
\usepackage{array}

\ifCLASSOPTIONcompsoc
  \usepackage[caption=false,font=footnotesize,labelfont=sf,textfont=sf]{subfig}
\else
  \usepackage[caption=false,font=footnotesize]{subfig}
\fi

\ifCLASSOPTIONcaptionsoff
  \usepackage[nomarkers]{endfloat}
 \let\MYoriglatexcaption\caption
 \renewcommand{\caption}[2][\relax]{\MYoriglatexcaption[#2]{#2}}
\fi
\begin{document}
%
\title{On the Effect of Observed Subject Biases \\in Apparent Personality Analysis \\from Audio-visual Signals}
%
%
%
%

\author{Ricardo~Dar\'io~P\'erez~Principi,
        Cristina~Palmero,
        Julio~C.~S.~Jacques~Junior,
        and~Sergio~Escalera
\IEEEcompsocitemizethanks{\IEEEcompsocthanksitem Ricardo~Dar\'io~P\'erez~Principi is a Master's student at Universitat de Barcelona, Spain, \protect 
E-mail: principidario@gmail.com}
\IEEEcompsocitemizethanks{\IEEEcompsocthanksitem Cristina Palmero is a PhD candidate at Universitat de Barcelona and Computer Vision Center, Spain, \protect
E-mail: crpalmec7@alumnes.ub.edu}
\IEEEcompsocitemizethanks{\IEEEcompsocthanksitem Julio C. S. Jacques Junior is a postdoctoral researcher at Universitat Oberta de Catalunya and research collaborator at Computer Vision Center, Spain, \protect
E-mail: jsilveira@uoc.edu}
\IEEEcompsocitemizethanks{\IEEEcompsocthanksitem Sergio Escalera is an associate professor at Universitat de Barcelona and Computer Vision Center, Spain, \protect
E-mail: sergio@maia.ub.es}
}
\IEEEtitleabstractindextext{%

\begin{abstract}
Personality perception is implicitly biased due to many subjective factors, such as cultural, social, contextual, gender and appearance. Approaches developed for automatic personality perception are not expected to predict the real personality of the target, but the personality external observers attributed to it. Hence, they have to deal with human bias, inherently transferred to the training data. However, bias analysis in personality computing is an almost unexplored area. In this work, we study different possible sources of bias affecting personality perception, including emotions from facial expressions, attractiveness, age, gender, and ethnicity, as well as their influence on prediction ability for apparent personality estimation. To this end, we propose a multi-modal deep neural network that combines raw audio and visual information alongside predictions of attribute-specific models to regress apparent personality. We also analyse spatio-temporal aggregation schemes and the effect of different time intervals on first impressions. We base our study on the ChaLearn First Impressions dataset, consisting of one-person conversational videos. Our model shows state-of-the-art results regressing apparent personality based on the Big-Five model. Furthermore, given the interpretability nature of our network design, we provide an incremental analysis on the impact of each possible source of bias on final network predictions.

\end{abstract}

}

\maketitle

\IEEEdisplaynontitleabstractindextext

%
\IEEEpeerreviewmaketitle

\IEEEraisesectionheading{\section{Introduction}\label{sec:introduction}}

%
%
%
%
Psychologists have developed reliable and useful methodologies for assessing personality traits~\cite{Gregory:2009}. However, personality assessment is not limited to psychologists, i.e., everybody, everyday, makes judgments about our personalities as well as of others. We are used to talk about an individual as being (non-)open-minded, (dis-)organised, too much/little focused on herself, etc.~\cite{Pianesi:ICMI:2008}. In the so-called ``first impressions'', people spontaneously attribute personality traits to unacquainted people in milliseconds, even from a still photograph, and quite consistently~\cite{Todorov:2017:book}.  Nonetheless, support for the validity of these impressions is inconclusive, raising the question of why do we form them so readily?

From a computational point of view, personality trait analysis, or simply personality computing, is studied from three main perspectives: automatic personality recognition, perception and synthesis~\cite{vinciarelli2014survey}. The first is related to the recognition of the \textit{real} personality of an individual, generally based on self-report questionnaire analysis. Personality synthesis, on the other hand, is associated to the generation of artificial personalities through embodied agents (e.g., applied in social interfaces or robotics). Personality perception, which is the focus of our work, is the research area centred on the analysis/recognition of personality others attribute to a given person, also referred to as apparent personality, or first impressions~\cite{ponce2016chalearn,escalante2018explaining}, although the latter covers a wide area and is not restricted to personality.


Personality computing has been receiving a lot of attention during the past few years~\cite{vinciarelli2014survey}. A recent comprehensive review on personality perception approaches~\cite{junior2018first}, centred on the visual analysis of humans, shows that most works developed for apparent personality recognition are combining standard machine learning methods with hand-crafted features, and that just few recent works have started to use more sophisticated solutions based on deep learning. The study shows that the state of the art in automatic personality perception is addressing the problem from different perspectives (e.g., during face-to-face interviews, conversational videos, small groups, human-computer interaction, etc.) and data modalities (e.g., using still images, image sequences, audio-visual or multi-modal). Furthermore, they report that deep neural networks are currently one of the most promising candidates to tackle the challenges of multi-modal data fusion on the topic.

The main concern about personality perception studies is that they are based on social/person perception. The central assumption behind such approach is that social perception technologies are not expected to predict the actual state of the target, but the state external observers attribute to it, i.e., apparent personality is conditioned on the observer. When machine learning based approaches are considered, these impressions are referred to as labelled data, used during development and evaluation stages. Nevertheless, the validity of such data can be very subjective due to several factors, such as cultural~\cite{Walker:2011}, social~\cite{Sutherland:2016}, contextual~\cite{Todorov2014}, gender~\cite{Mattarozzi:PLOS:2015} or appearance~\cite{Rudert2017101}. Such biases, present in current datasets, make research on personality perception a very challenging task.

In this work, we study different sources of bias affecting personality perception, as well as the impact they have on prediction ability. Such biases are analysed incrementally so that their impact can be measured when compared to a baseline. Note that our goal is not to automatically identify such biases. Rather than this, we exploit their influence to improve the recognition performance of apparent personality. To do so, we propose a deep neural network based solution to simultaneously deal with spatio-temporal information, data fusion and subjective bias for automatic personality perception in one-person conversational videos. Our study is motivated by the fact that dedicated (task-specific) networks can be used to automatically recognise different attributes\footnote{In this work, we use \textit{attributes} and \textit{factors} interchangeably.} of people in the images, being able to provide some explanation of the results as well as to improve overall accuracy performance. Hence, we benefit from state-of-the-art deep learning architectures and data fusion strategies. Furthermore, we perform spatio-temporal feature analysis to somehow address the recurrent problem on the topic related to \textit{slice size/location} (detailed in Sec.~\ref{sec:slices}). As far as we know, no existing work on the topic analyses the influence of \textit{slice size/location} together with subjective bias.

We base our study on the ChaLearn First Impressions (FI) dataset~\cite{ponce2016chalearn}, which is currently the largest, public and labelled dataset on the field, composed of 10K short video clips of individuals talking to a camera. Although different trait models have been proposed and broadly studied over the past decades, we focus on the recognition of Big-Five~\cite{McCrae:1992} traits, as it is one of the most adopted models in psychology and personality computing. It is also known as Five-Factor Model, often represented by the acronym OCEAN, which is associated with the five following dimensions: \textit{\textbf{O}penness to Experience}, \textit{\textbf{C}onscientiousness}, \textit{\textbf{E}xtraversion}, \textit{\textbf{A}greeableness}, and \textit{\textbf{N}euroticism}. 

In summary, our main contributions are as follows:
\begin{itemize}
\item We employ different task-specific networks to both extract high-level information from data and to analyse human bias affecting personality perception with respect to: 1)~facial emotion expressions; 2)~attractiveness; 3)~age; 4)~gender; and 5)~ethnicity of observed subjects\footnote{Attribute categories used in this research are imperfect for many reasons. For example, there is no gold standard for ``ethnicity`` categories, and it is unclear how many race and gender categories should be stipulated (or whether they should be treated as discrete categories at all). This work is based on an ethical and legal setting, and the methodology and bias findings are expected to be applied later to any re-defined and/or extended attribute category.}.
\item Audio-visual data are combined with the outputs of those task-specific networks using a late fusion strategy before regressing the Big-Five traits at test stage. Then, the influence of different attributes is incrementally analysed and discussed. This way, the subjectivity associated with analysed attributes can be partially explained, which is aligned with recent studies on the topic of explainability/interpretability~\cite{escalante2018explaining} in machine learning.
\item Different ways to select and represent temporal information are analysed, as well as their effect on apparent personality recognition. Thus, the influence of subjective biases is also investigated on the temporal domain.

\item We improve the state of the art on the adopted dataset and demonstrate that complementary information (i.e., acoustic, visual, spatio-temporal and high-level attributes represented by deep features) can be used to boost accuracy performance. More concretely, we found the best performance by the combination of raw images, audio, age, facial expressions, and attractiveness.
\end{itemize}

The remainder of the paper is organised as follows. Sec.~\ref{sec:related} presents the state of the art in automatic personality perception, with particular focus on deep learning solutions. The proposed architecture is detailed in Sec.~\ref{sec:model}. Experimental analyses are discussed in Sec.~\ref{sec:results}. Final remarks are drawn in Sec.~\ref{sec:conclusions}.

\section{Related work}\label{sec:related}


In this section, we present the state of the art on the research topic. Relevant works addressing automatic personality perception are discussed in Sec.~\ref{sec:app}. Sec.~\ref{sec:slices} describes approaches analysing \textit{slice size/location} in first impression studies. Then, a brief review about subjective bias analysis is presented in Sec.~\ref{sec:bias}. We refer the reader to~\cite{junior2018first} for a comprehensive review on the topic of apparent personality trait analysis, which includes an extensive discussion about subjectivity in data labelling from first impressions. 

\subsection{First impressions of personality}\label{sec:app}
Faces have been considered a rich source of cues for apparent personality or social traits perception in interpersonal impressions. Works relying on still image-based analysis have mostly focused on hand-crafted facial features followed by standard machine learning methods~\cite{al2014face} and, more recently, deep learning \cite{dhall2016first}. For instance, Vernon et al.~\cite{vernon2014modeling} employed geometrical and appearance facial attributes to model social factor dimensions. They observed that the mouth shape is positively related to \textit{approachability}, eyes geometry to \textit{youthful-attractiveness}, and masculine appearance to \textit{dominance}. Similarly, Guntuku et al.~\cite{guntuku2015others} used whole-image low-level features (e.g., colour histograms and texture analysis) to detect mid-level cues such as gender, age, emotional positivity, and eyes looking at camera, which in turn were used to predict real and apparent personality traits. Their results show that emotional positivity is significantly correlated with all traits, while eyes looking at camera is positively correlated to \textit{Agreeableness} and \textit{Openness} in perceived personality. 

When temporal information is available (e.g., image sequences), works can benefit from scene dynamics, acoustic information (if audio-visual data is provided), verbal content analysis or even data acquired by more sophisticated sensors (in the case of multi-modal)~\cite{junior2018first}. Each additional cue may bring useful and complementary information.


\subsubsection{Hand-crafted features}
Early works from Biel et al.~\cite{biel2012facetube,biel2013youtube} on the topic found \textit{Extraversion}
to be the easiest trait to judge by external observers and the most reliably predicted using audio and visual cues, with the largest cue utilisation, followed by \textit{Openness to Experience}. In \cite{biel2012facetube}, they exploited different cues: acoustic, such as speaking activity descriptors and prosodic features (e.g., energy and fundamental frequency); from face, looking activity (frontal face detection as a proxy for looking-at-camera events) and proximity from camera; and finally, from visual cues, the overall motion by means of weighted Motion Energy Images (wMEI). Co-occurrence events were also analysed, such as looking while speaking. The authors performed correlation analysis among audio and visual cues with Big-Five personality traits. Results show a wide variety of correlations, suggesting that each personality trait can be better predicted when a particular cue (or feature set) is exploited. In~\cite{biel2013youtube}, they added facial expressions (Ekman emotions) to predict personality, based on aggregated statistics and activity patterns of frame-by-frame predictions. Results show that emotion-based cues outperformed overall visual activity. In a similar study, Teijeiro et al.~\cite{teijeiro2014your} found that, following previous evidence, only \textit{Extraversion} could be predicted with statistical significance, and that among all sets of analysed features, statistics of emotion activity patterns were the most predictive cue. Furthermore, their experiments revealed that, at least for this trait, the first seconds of a video predicted the observers' impressions better, which is aligned with recent studies showing that first impressions are build as brief as a blink of an eye~\cite{Todorov:2017:book}. However, as recently demonstrated by Gatica-Perez et al.~\cite{gatica2018vlogging} relying on the same feature set, the use of multiple videos of the same user helps achieve higher performance in apparent personality inference. This suggests that exposure variability allows for a more robust and reliable personality estimation, and also adds to the fact that personality remains stable over time. 

Despite the great advances of previous work~\cite{biel2012facetube, biel2013youtube,teijeiro2014your} to push the research on the topic, their studies are limited to analyses performed on relatively small datasets (i.e., composed of \texttildelow450 short video clips), which does not guarantee generalisation to different target populations. In~\cite{gatica2018vlogging}, a longitudinal dataset was introduced to study social impressions. Although a total of 2,376 videos were used, they were collected from just 99 YouTube users.

\subsubsection{Deep-based methods}
Following the advent of deep learning in computer vision and machine learning, the tendency has shifted from hand-crafted features towards raw inputs and end-to-end learning systems, even though aggregation of statistics over time still predominates. 

G\"{u}rpinar et al. \cite{gurpinar2016combining} proposed a pre-trained two-stream Convolutional Neural Network (CNN) to extract facial expressions and scene information, and then combined their features to feed a kernel Extreme Learning Machine (ELM) regressor. Facial features are represented by computed statistics over the sequence of frames, whereas scene features are extracted from the first frame of the video. 
The fusion of the two modalities proved to be beneficial for all traits except for \textit{Neuroticism}, where simple Local Gabor Binary Patterns from Three Orthogonal Planes (LGBP-TOP) achieved the best performance. Later, they added acoustic features and investigated different fusion approaches~\cite{gurpinar2016multimodal}. The best results were obtained with a weighted score level fusion of acoustic, deep and LGBP-TOP features.

To further leverage temporal information, Subramaniam et al. \cite{subramaniam2016bi} proposed to segment audio and video in fixed partitions so that inter-partition variations are learned as temporal patterns. Video is further pre-processed by selecting one face per partition. Two end-to-end approaches are proposed. Each network is composed of two branches, one for encoding audio and the other for visual features. The first model, formulated as a 3DCNN network (which has the capability of exploiting short and local temporal information), combines visual (i.e., temporally-ordered aligned faces) and acoustic information using a late fusion strategy. The second architecture jointly models each face and audio partitions using CNN and Fully Connected (FC) layer, respectively. Then, resulting features per partition are combined and fed into a Long Short-Term Memory (LSTM) module, which generates a prediction from each time step. Finally, the individual scores are averaged to output a final prediction. The latter approach obtained superior performance, presumably due to the use of both audio and visual features to jointly learn the temporal correspondences. 

G\"{u}\c{c}l\"{u}t\"{u}rk et al.~\cite{guccluturk2016deep} presented an end-to-end two-stream deep residual network that uses raw audio and video with feature-level fusion which, contrary to aforementioned works, does not require any feature engineering (such as face detection and alignment). The study was extended~\cite{guccluturk2017multimodal} to investigate the combination of different modalities, including verbal content analysis. As expected, the fusion of all modalities outperformed audio-visual fusion, followed by visual, audio, and language features individually. 
Ventura et al.~\cite{ventura2017interpreting} further investigated why CNN models obtain such impressive accuracies on personality perception. They found that the internal model representations mainly focused on the speaker's face and, more specifically, on key semantic regions such as mouth, eyes, eyebrows, and nose. 

Concurrently, Wei et al. \cite{wei2018deep} proposed a Deep Bimodal Regression framework, where Big-Five traits are predicted by averaging the scores obtained from different models. They modified the traditional CNN to exploit important visual cues (introducing what they called Descriptor Aggregation Networks), and built a linear regressor for the audio modality. To combine complementary information from the two modalities, they ensembled predicted regression scores by both early and late fusion. In their work, the best performance was obtained through the combination of both modalities, reinforcing the benefits of feature fusion.
%
%

More recently, Bertero et al. \cite{kampman2018investigating} combined raw audio, one random frame per video and verbal content in a tri-modal stacked CNN architecture. Different fusion strategies were analysed, i.e., decision-level fusion via voting (as a linear combination of weights per modality and trait), feature-level fusion, end-to-end and last-layer training. Results show that end-to-end training performed significantly better for all traits except \textit{Agreeableness}, which was harder to predict for all fusion alternatives. 

The great majority of deep learning-based works~\cite{gurpinar2016combining,gurpinar2016multimodal,subramaniam2016bi,guccluturk2016deep,guccluturk2017multimodal,ventura2017interpreting} adopted the FI~\cite{ponce2016chalearn} dataset, 
which is, to date, the largest publicly available dataset on the topic, suitable to study apparent personality by deep learning strategies. 
As it can be observed, disregarding the different model architectures, feature representations and fusion strategies, no existing work has proposed to analyse and ``explain'' the subjective bias in personality perception, in particular together with \textit{slice size/length} analysis. As opposed to previous works, we propose to simultaneously advance the research on the field while performing an incremental analysis of different attributes affecting apparent personality recognition. 

\subsection{Slice size/locations}\label{sec:slices}

When temporal information is considered, the decision of which part of the video (from the whole sequence) will be analysed is a common requirement to be addressed. Usually known as \textit{slice size/location}, it affects the recognition performance and, in most of the cases, is empirically defined. Although G\"u\c{c}l\"ut\"urk  et al.~\cite{guccluturk2017multimodal} show that there is enough information about personality in a single frame, when evaluating the changes in performance as a function of exposure time, according to the authors, the same reasoning does not hold for the auditory modality (especially for very short auditory clips). Furthermore, the standard approach for ``frame selection'' is based on random choices or uniform distributions.

According to~\cite{Nguyen:epfl:thesis:2015}, one of the challenges in thin slice research (i.e., when just a small fraction of time is observed, usually from few seconds to \texttildelow5 minutes) is the amount of temporal support necessary for each behavioural feature to be predictive of the outcome of the full interaction. In other words, some cues require to be aggregated over a longer period than others. Although different works have been proposed to analyse the effect of \textit{slice size/location} on the accuracy performance~\cite{teijeiro2014your,Lepri:TAC2012}, no metric assessing the necessary amount of temporal support for a given feature exists~\cite{Nguyen:epfl:thesis:2015,junior2018first}.

\subsection{Subjective bias in first impressions}\label{sec:bias}

Subjective bias in personality perception is an emergent research topic~\cite{junior2018first}. First impressions have been analysed in psychological studies and several factors influencing/affecting person perception have been reported, being  cultural~\cite{Walker:2011}, social~\cite{Sutherland:2016}, contextual~\cite{Todorov2014}, gender~\cite{Mattarozzi:PLOS:2015} and appearance~\cite{Rudert2017101} some of them. In this section, we briefly discuss relevant studies analysing subjective bias in first impressions, in particular those related to the attributes analysed in our work. Note that attentive explanations models~\cite{Park:CVPR:2018} based on textual/multi-modal justifications could be used to localise the evidences that support personality perception analysis. However, there is neither standard protocol nor design in personality computing for such data annotation procedure, which makes such task a big challenge. Subjective bias analysis is an ample topic, probably unbound, and a comprehensive review goes beyond the scope of this study.   

\textbf{Facial emotion expression.} According to~\cite{sutherland2017facial}, the most important
source of within-person variability, related to social impressions of key traits of \textit{trustworthiness}, \textit{dominance}, and \textit{attractiveness}, is the emotional expression of the face. A few works found in the literature propose to analyse first impressions using synthetic~\cite{Todorov:cognition} or manipulated faces~\cite{walker2011universals}, so that different face attributes can be controlled, such as facial expressions. From the computing point of view, facial expression recognition is one of the most studied topics in the computer vision community, for which a wide number of solutions exist~\cite{corneanu2016survey}.

\textbf{Attractiveness.} The effects of appearance on social outcomes may be partly attributed to the \textit{halo effect}~\cite{Todorov:book:2012}, which is the tendency to use global evaluations to make judgments about specific traits (e.g., \textit{attractiveness} correlates with perceptions of \textit{intelligence}), having a strong influence on how people build their first impressions about others~\cite{lorenzo2010beautiful}. Perceived beauty is certainly a subjective and abstract factor that strongly depends on the observer's bias, particularly linked to age and gender of both observer and target~\cite{foos2011adult}.  
Attractiveness measured from faces has been long analysed on the computing domain from hand-crafted geometric and texture features \cite{chen2010benchmark}. Nonetheless, end-to-end methods are slowly emerging with greater performance, as well as the development of large and public datasets to push the research on the topic~\cite{liang2018scut}.

\textbf{Age.} Evidences found in the literature show that age may be linked to differences in perceived personality. For instance, Chan et al.~\cite{chan2012stereotypes} reported that \textit{Neuroticism}, \textit{Extraversion} and \textit{Openness} are perceived to be highest in adolescents, while \textit{Agreeableness} is highest among the elderly. However, its effect on automatic personality perception has not been studied in depth yet. Escalante et al.~\cite{escalante2018explaining} analysed the effect of age perception when studying first impressions in the context of job recommendation systems. Results show that, although non-consciously, people prefer to invite for a job interview women when they are younger and men when they are older, which reflects a latent bias towards age and gender. Although age estimation has been receiving a lot of attention from the computer vision community, it still remains a challenging task due to \textit{``the uncontrolled nature of the ageing process''}~\cite{agustsson2017apparent}.%

\textbf{Gender.}\label{sec:sub_gender} This is another factor known to interact not only with personality judgments, commonly due to stereotypes as in the case of age, but also in \textit{real} personality across cultures~\cite{costa2001gender}. For instance, it was reported in~\cite{escalante2018explaining} that women tend to be viewed as more open and extroverted, compared to men. The authors also observed an overall positive attitude/preconception towards females in both personality traits (except for \textit{Agreeableness}) and analysed job interview variable. As it can be seen, the subjectivity associated to gender can have a strong impact on first impressions. Recent studies show that Artificial Intelligence (AI) based models are exhibiting racial and gender biases~\cite{escalante2018explaining,junior2018first}, which are extremely complex and emerging issues. Even though it may be unintentional and subconscious, the ground truth annotations used to train AI based models may reflect the preconception of the annotators towards the person in the images or videos. From a computer vision point of view, state-of-the-art approaches generally combine gender with age~\cite{levi2015age} or ethnicity~\cite{narang2016gender} in a multi-task framework, or even consider all these problems jointly~\cite{guo2014framework,han2018heterogeneous}. 

\textbf{Ethnicity.} The latent bias towards apparent ethnicity in first impressions has been recently investigated in~\cite{escalante2018explaining}, whose results indicated an overall positive bias towards Caucasians, a negative bias towards African-Americans, and no discernible bias towards Asians. Moreover, gender bias was observed to be stronger compared to ethnicity bias. To the best of our knowledge, computer vision/machine learning literature on ethnicity estimation as an individual task is scarce~\cite{hosoi2004ethnicity}. Instead, it is usually combined with other attributes as in the case of gender~\cite{narang2016gender,han2018heterogeneous}.

In general, to isolate and analyse the effect/influence of each particular attribute with respect to first impressions is not an easy task, as they are strongly correlated and people probably build their impressions of others taking into account all attributes together. However, as it is described in the next sections, in this study we analyse the effect of each above-mentioned attribute on the accuracy performance of automatic personality perception. In particular, we study how the recognition of a particular attribute can boost automatic personality perception.



\section{Proposed Model}\label{sec:model}
\label{sec:method}

\begin{figure*}[htbp]
	\centering
	\includegraphics[width=0.9\textwidth]{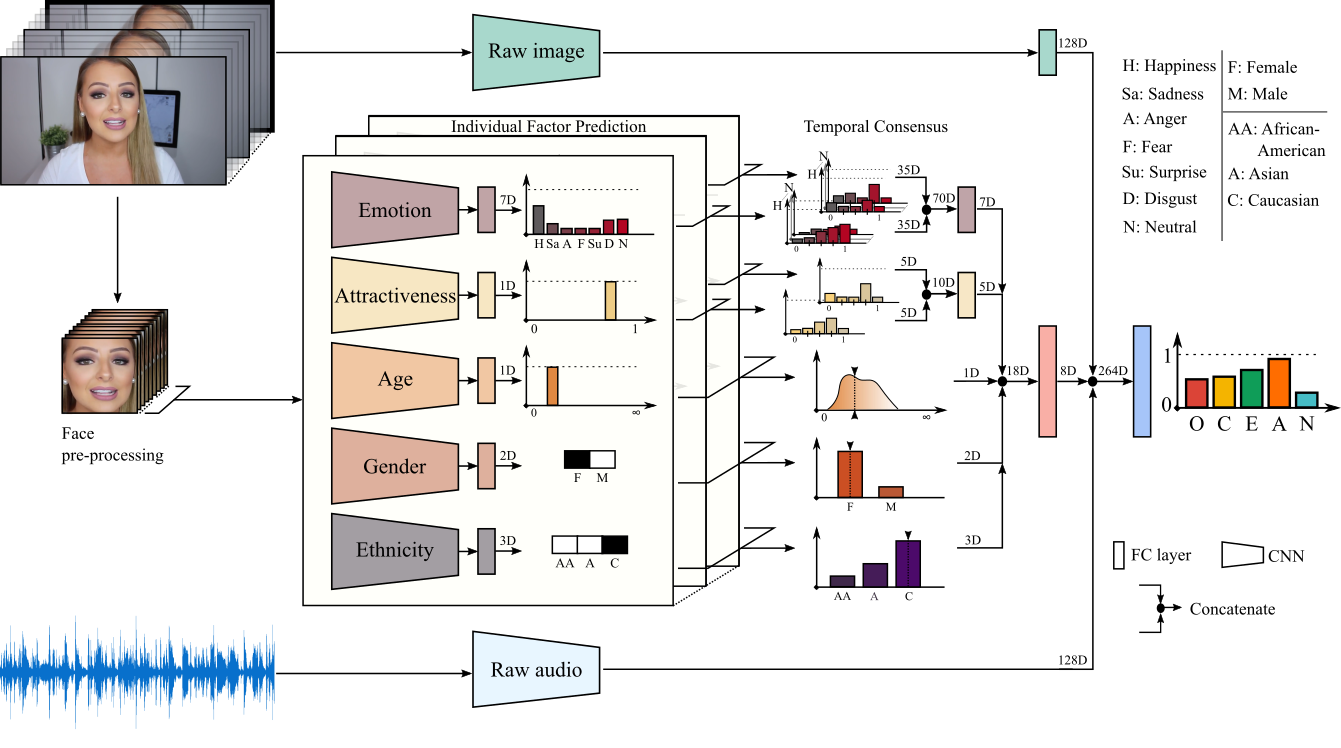}
	\vspace{-0.2cm}
	\caption{Overview of the proposed method. For each video, task-specific models are used to predict per-frame estimates of each attribute, which are then aggregated to obtain a video-level prediction using different spatio-temporal aggregation approaches. Next, we combine this representation of attributes with raw audio (obtained from the whole video) and scene descriptors / visual information (i.e., ``Raw image'': features extracted from a selected frame of the video), and jointly model them by late fusion to regress the Big-Five personality traits.}
	\label{fig:pipeline}
\end{figure*}

In this work, we propose a multi-modal deep neural network that combines features from the raw visual and audio streams along with the predicted values of different attributes, associated to the observed subjects, to predict apparent Big-Five personality traits. To partially consider the subjective bias in first impressions, we examine facial expressions, attractiveness, age, gender and ethnicity of observed people, which are automatically estimated based on task-specific networks. 
This way, instead of relying just in the raw input, we guide the network with explicit information that may help in the final personality prediction. This, in turn, provides mechanisms to analyse the influence of each possible bias factor of the observed subject in relation to his/her overall impression. 
An overview of the proposed model is shown in Figure \ref{fig:pipeline}. In summary, our approach consists of four main stages:
\begin{enumerate}
    \item \textbf{Individual factor prediction.} We use pre-computed models to recognise the different attributes (emotion, attractiveness, age, gender and ethnicity). Since performance is strongly affected by the choice of architecture and hyperparameters, we select state-of-the-art methods trained and validated in their respective tasks. 
    \item \textbf{Temporal consensus.} Considering the whole set of frames per video, per-frame individual attribute estimates are aggregated to obtain video-level predictions of each attribute. Moreover, dynamic attributes such as facial emotion expression and attractiveness are encoded using different video segments.
    \item \textbf{Modality fusion.} Visual, acoustic and high-level attributes are represented and combined in different ways (see Table~\ref{tab:sources} and Figure~\ref{fig:pipeline}). Each training/test sample is defined by a selected frame of the video, the whole raw acoustic waveform and the previously-estimated video-level attributes. In particular, we select $M$ uniformly-distributed raw images from the whole video, thus having $M$ samples per video. Aggregated attributes are jointly modelled to produce additional-contribution features. Finally, all information is combined in a late fusion scheme for personality trait regression.
    \item \textbf{Video-level personality traits prediction.} At test stage, the final video-level prediction per trait is computed as the median of the $M$ individual personality traits predictions of a video.
    
\end{enumerate}

Next, we describe a generic model that combines all attributes and modalities. However, in Sec. \ref{sec:results} we test different combinations of attributes and modalities, thus producing different model architectures and sizes of fusion feature vectors.

\subsection{Raw modalities}

\subsubsection{Visual}
Raw image sequences are one of the most informative sources in apparent personality recognition~\cite{junior2018first}. However, contiguous frames contain redundant information, in particular when high frame rates are considered. Traditionally, single images have been used as an alternative, thus decreasing the computational burden associated to analysing an entire video. Therefore, we select a subset of $M$ equidistant video frames to have representative samples of visual cues, which has been shown to be highly discriminant and computationally efficient~\cite{zhang2016deep}. 
In our approach, we treat each frame as an independent sample, such that all selected frames of the same video have the same associated labels with respect to Big-Five personality traits. Each frame is resized to $224$ x $224$ pixels and fed to a ResNet-50 network pre-trained with Imagenet weights, followed by a final FC layer to reduce dimensionality, thus producing a 128D feature vector (illustrated in the upper part of Figure~\ref{fig:pipeline}). 
This way, instead of just focusing on the face region (as described in Sec.~\ref{sec:explicit}), we also consider the person's body, background and overall scene, which have demonstrated to be rich sources of information to code personality~\cite{dhall2016first}.

\subsubsection{Audio}
The auditory modality is a complementary source of information that has proved to be beneficial in emotion and personality prediction \cite{junior2018first}. In this work, we follow the approach of G\"{u}\c{c}l\"{u}t\"{u}rk et al. \cite{guccluturk2016deep} and use the raw waveform of the whole video as input to a modified ResNet-18 in which convolutional, pooling kernels, and strides are one-dimensional to adequate to the audio dimensionality. We remove the last convolutional block of the modified network, resulting in a 14-layer network, and add an average pooling layer to output a final 128D feature vector (illustrated in the lower part of Figure~\ref{fig:pipeline}).

\subsection{Individual factor sources: high-level attributes}
\label{sec:explicit}


All additional factors included in our approach concentrate on the speaker's face. Therefore, a pre-processing step is applied to all video frames to restrict the network's attention and increase generalisation ability. In particular, faces are detected and extracted with a HOG-SVM face detector \cite{dlib09}, followed by rigid face alignment using a 68-landmark model \cite{kazemi2014one}, resulting in a $224$ x $224$ face image. It is worth considering that alignment may remove pose information; however, head pose is not relevant for the studied set of attributes, and it is already implicit in the raw video frame.  Frames with no detected faces are discarded. For each factor, all detected faces in a video are individually processed and their inferred scores are combined (detailed next) with specific aggregation functions to reach a video-level consensus, thus capturing variability over time instead of relying on instantaneous estimates. 

Some of the studied attributes remain constant throughout the whole video, namely age, gender, and ethnicity. However, facial expressions and attractiveness are treated as dynamic factors. Therefore, for the latter we also study their temporal evolution. Factor-specific details are summarised in Table \ref{tab:sources}. 

 {\renewcommand{\arraystretch}{1.3} 
\begin{table*}[!t]
\centering
\caption{Description of the information sources included in our study.}
\begin{tabular}{C{1.8cm}|C{1.8cm}|C{2.8cm}|C{6.0cm}|C{3.0cm}}

\hline
\textbf{Information source}  & \textbf{Network}  & \textbf{Dataset} & \textbf{Temporal aggregation} & \textbf{Output feature dimension}       \\ \hline
Visual     & Resnet50 + FC(128) & First Impressions~\cite{ponce2016chalearn} & None    & 128D \\ \hline
Age       & WideResNet         & IMDB-WIKI~\cite{rothe2015dex}         & Median & 1D  \\ \hline
Gender    & VGGFace              & IMDB-WIKI~\cite{rothe2015dex}         & Mode    & 2D (one-hot encoding) \\ \hline
Ethnicity & VGGFace              & UTKFace~\cite{zhifei2017cvpr}           & Mode    & 3D (one-hot encoding) \\ \hline
Attractiveness$^*$ & VGGNet         & SCUT-FBP5500~\cite{liang2018scut} & \makecell{5-bin histogram for the whole video, or \\ 5-bin histogram for each segment\\}           & 5D                  \\ \hline
Facial expression$^*$ & AlexNet  & Ensemble~\cite{rosa2018deep}   &  \makecell{5-bin histogram per emotion for whole video, or \\  5-bin histogram per emotion for each segment\\}   & 7D    \\ \hline
Audio             & Modified ResNet18               & First Impressions~\cite{ponce2016chalearn} & None          & 128D                   \\ \hline
\multicolumn{5} {l} {$^*$ \textit{Dynamic} attribute: it can be represented in different ways on the temporal domain.}
\end{tabular}
\label{tab:sources}
\end{table*}
}

\subsubsection{Emotions}\label{sub:emotion}

According to the literature, facial expressions have been extensively used for personality trait analysis~\cite{Todorov:book:2012,junior2018first}. In this work, we rely on the study of Rosa et al.~\cite{rosa2018deep}, which compared AlexNet, ResNet and VGG16 trained on an ensemble of datasets, 
selected to maximise ethnicity, age, overall appearance and scenario variability. Approximately 37,800 images were collected in their study. The authors also applied data augmentation to balance the resulting dataset. 
Despite ResNet slightly outperformed other architectures in their study, we selected AlexNet for our approach as it offers a trade-off between accuracy and training speed. 

In our work, facial expressions are classified into one of the 6 Ekman basic emotions (\textit{Anger}, \textit{Disgust}, \textit{Fear}, \textit{Happy}, \textit{Sadness}, and \textit{Surprise}), with the addition of the \textit{Neutral} expression, using a softmax layer. We use the output probabilities distribution as frame-level predictions and aggregate them into a video-level descriptor by means of a normalised 5-bin histogram per emotion, which results in a 35D feature vector. Additionally, to analyse the effect that the order in which emotions are expressed has on perceived personality, we divide the video in two equal-length partitions and perform a partition-wise histogram aggregation, thus resulting in a 70D feature vector. Regardless of the aggregation method, the output feature vector is reduced to a 7D final vector by means of a FC layer with ReLU activation.  

\subsubsection{Attractiveness}


To obtain attractiveness values of people in the FI dataset~\cite{ponce2016chalearn}, we fine-tune a VGG16 model (pre-trained on ImageNet) on the recently released SCUT-FBP5500 dataset~\cite{liang2018scut}. This dataset consists of 5,500 frontal images with age and gender variability, including 4,000 Asian and 1,500 Caucasian subjects with 50\% males and 50\% females in both sets. It is important to note that the dataset was labelled by volunteers aged 18-27 years, which is likely to impact the ground truth, as young adults tend to rate young faces as more attractive than old ones \cite{foos2011adult}. To regress the attractiveness value, we simply replace the last layer of the original VGG16 model by a 1D FC regression layer. The network is trained in two steps. First, training only the last FC layers, and then fine-tuning the whole model. The attractive scores on the SCUT-FBP5500 dataset range between $[1,5]$; however, we scale them using a linear function to range between $[0,1]$. 

Perceived beauty can change over time depending on the face angle, facial expression, scene dynamics, etc. In order to capture this variation throughout a video, we compute a normalised 5-bin histogram from the face-level scores, producing a 5D feature vector. Similarly as in the case of emotion (Sec.~\ref{sub:emotion}), we also analyse the influence of changes in perceived beauty over time by segmenting the video in two fragments of same length and computing a histogram for each one, thus producing a 10D feature vector. If using the later approach, a FC layer with ReLU non-linearity is added on top to convert the 10D feature vector into a 5D one, so as to maintain an equilibrium among the number of features of each factor.

\subsubsection{Age}


Our age estimation model is based on the work of Rothe et al.~\cite{rothe2015dex}. However, we use a WideResNet trained from scratch to reproduce their results, adding a linear regression layer on top to predict the age value per face image. The model is fine-tuned on the IMDB-WIKI dataset~\cite{rothe2015dex}, the largest publicly available dataset of face images including gender and age labels, consisting of a total of 524,230  images. Since age is a stable factor throughout a short video, we take the median value of the individual scores to obtain a video-level representation.

\subsubsection{Gender}\label{sub:gender}


Due to the scarcity of approaches tackling gender estimation as a single task problem, we modify a VGG16 pre-trained with Faces \cite{Parkhi15} and fine-tune it on the IMDB-WIKI~\cite{rothe2015dex} dataset to predict gender. In our work, gender is represented using one-hot encoding, as it is a binary categorical variable (\textit{Female} and \textit{Male}), resulting in a 2D vector. Since gender is a static factor, we use a voting aggregation scheme to decide the predicted video label. 

\subsubsection{Ethnicity}


Automatic ethnicity recognition is generally tackled as a multi-task problem, as in the case of gender. Thus, we follow a similar approach chosen for gender estimation to predict ethnicity labels. However, in this case we modify the final layer of the same VGG16 model used above to predict three categories (\textit{African-American}, \textit{Asian}, and \textit{Caucasian}), and fine-tune it on the UTKFace~\cite{zhifei2017cvpr} dataset. This dataset consists of over 20,000 face images covering a large variation in facial expression, pose, occlusions and illumination conditions. Ethnicity labels are originally comprised of \textit{White}, \textit{Black}, \textit{Asian}, \textit{Indian}, and \textit{Others} (i.e., \textit{Hispanic}, \textit{Latino} and \textit{Middle Eastern})\footnote{\url{https://susanqq.github.io/UTKFace}}. 
To adapt such categorisation (and dataset) to our needs, we discarded images with \textit{Indian} labels, as such category is not included in the FI~\cite{ponce2016chalearn} dataset. We further grouped \textit{Others} and \textit{White} categories, and considered them to belong to the \textit{Caucasian} group. Finally, labels are one-hot encoded, producing a 3D feature vector per face image. In the same way as gender, individual scores for ethnicity are aggregated by computing their mode.

\subsection{Fusion of modalities}

Fusion of modalities is performed in two steps (illustrated in Figure~\ref{fig:pipeline}). First, the video-level attribute predictions are concatenated in a 18D feature vector to model their joint contribution by means of a FC layer, which reduces its dimensionality to 8D. Second, such 8D feature vector is combined along with the raw audio (128D) and visual features (128D), producing a 264D feature vector, which is fed to the final FC layer with sigmoid activation and 5 units, responsible for regressing the Big-Five personality traits in the range $[0,1]$. It should be noted that the individual factors can be viewed as complementary information, having smaller impact on the results if compared to audio-visual information, which is reflected by the small size of their output vectors. Nevertheless, while having small impact on the outcomes, they have strong explanation capabilities, as shown in Sec.~\ref{sec:attributes:analysis}.
%


\subsection{Training strategy}

Learning is carried out in a stage-wise fashion. First, all task-specific networks responsible for recognising the high-level attributes are individually trained on their respective datasets. Then, the frame-based and video-level factor predictions are computed on the FI training set (described in Section \ref{sec:dataset}). Third, for each selected frame of a video, we extract the scene descriptor (i.e., represented by ``raw image'' in Figure~\ref{fig:pipeline}) from a frozen visual stream, while the audio stream, all FC layers of the temporal consensus module, as well as the remaining layers are trained from scratch for 40 epochs with a learning rate of 0.001. Note that temporal information is somehow encoded in the audio and \textit{dynamic} attributes (attractiveness and facial expressions) based on their representations. Finally, we fine-tune the whole model, except for the independent factor models, with a learning rate of 0.0005 during 100 epochs. To improve the learning process, the learning rate was automatically adjusted by a factor of 0.95 when no improvement was detected after 5 epochs. We used Mean Absolute Error (MAE) on the validation set as metric to decide when to reduce it, and Mean Squared Error (MSE) as training loss. Training was performed using ADAM optimiser with a batch size of 25 frames. $M$ is set to $10$ frames per video for this stage.

At test stage, $M$ is set to $50$ and the final video-level prediction is defined by the median prediction per trait, taking into account the $M$ frame-based estimations.

\section{Experiments}\label{sec:results}
In this section, we evaluate the effectiveness of the proposed method on the FI~\cite{ponce2016chalearn} dataset, and compare with state-of-the-art audio-visual based methods. Furthermore, we analyse the effect that each analysed source of information has on the recognition performance as well as on personality perception. We follow the evaluation protocol used in the ChaLearnLAP First Impressions challenge \cite{escalante2018explaining}. We report the average accuracy for each of the 5 traits individually as:

\begin{equation}\label{accuracy_eqn}
	acc_{j}=1 - \frac{\sum_{i=1}^{N} \abs{p_{ij} - gt_{ij}}}{N},
\end{equation}
%
where $p_{ij}$ is the predicted value for video $i$ and trait $j$, $gt_{ij}$ their respective ground truth value, and $N$ the number of videos in the test set. 

Note that, as shown in~\cite{escalante2018explaining}, the ground truth values of the training and test partitions of the FI dataset follow a very similar normal distribution, with most values concentrated in a small fraction (i.e., at the centre) of the whole distribution for every Big-Five trait. As a matter of fact, the authors demonstrated that just by using the average value of each trait from the training set, they could reach around 88\% accuracy on the test set. Thus, a small improvement in the obtained accuracy means a high increase in model generalisation capability.


\subsection{First Impressions (FI) dataset}
\label{sec:dataset}
The First Impressions~\cite{ponce2016chalearn} dataset was released at ECCV 2016 Challenge. It consists of 10,000 small video clips of approximately 15 seconds each, extracted from more than 3,000 creative commons YouTube high-definition videos of people facing and speaking in English to a camera. Participants show gender, age, nationality, and ethnicity variability. Videos include only one person, the scene is clear and speakers face front at least 80\% of the time. Videos were labelled with personality trait variables by zero-acquaintance raters using Amazon Mechanical Turk (AMT) through pair-wise ranking annotations, which were then transformed into continuous values by fitting a Bradley-Terry-Luce with maximum likelihood. Annotated traits correspond to the Big-Five model, providing a ground truth value for each of the 5 traits within the range $[0, 1]$. The dataset was split into train, validation and test sets following a 3:1:1 ratio. Note that some individuals may appear in different sets due to the video split procedure, but at different time intervals. However, since labels were taken per video, the same person may have different ground truth values in each video. 

\subsection{Performance of the individual factor models}\label{sec:individual:models}
The performance of the proposed model strongly depends on the robustness and accuracy of the additional information source models. For this reason, we first evaluate them on the test sets of their respective datasets used for training. Those datasets that do not have a specific test set were split following a 9:1 ratio. 
To do so, we considered gender and ethnicity labels provided with the dataset~\cite{ponce2016chalearn}, and manually annotated a subset of the test set with emotion labels. In particular, for each emotion, we first selected all the images for which the predicted score of that emotion was over a threshold of $0.5$. Then, such images were manually annotated in a random order, until a balanced number of images per emotion was reached (approx. 285 images), using just 1 image per video. Note that we did not evaluate attractiveness and age attributes on the FI dataset as these variables are too subjective to be manually annotated in the scope of this work. 

Results are summarised in Table~\ref{tab:fac:models}. 
The obtained performances for all the tasks are on par with the state of the art, and hence acceptable for the purpose of our approach. It should be noted that the emotion recognition model was trained on facial expressions in absence of speech, so we would expect a decline in performance when applied to the talking scenario the FI dataset portrays, containing sudden changes of poses and facial expressions altered by speech. Nevertheless, the model demonstrated to reach competitive accuracy.


{\renewcommand{\arraystretch}{1.3}
\begin{table}[t!]
\caption{Recognition performance of factor models on their respective databases and on the First Impressions (FI) test set. Metrics: F1 score for emotion, gender and ethnicity (classification task); and MAE for age and attractiveness (continuous ordered values).}
\begin{tabular}{c|c|c|c|c|c}
\hline
\textbf{Dataset} & \textbf{Emotion} & \textbf{Gender} & \textbf{Ethnicity} & \textbf{Age} & \textbf{Attract.} \\ \hline
See Table~\ref{tab:sources} & 0.96 &  0.93 & 0.94  & 0.22   & 0.23     \\ \hline
FI~\cite{ponce2016chalearn} & 0.85 & 0.90  & 0.90  & -     & -  \\ \hline
\end{tabular}
\label{tab:fac:models}
\end{table}
}

\subsection{Effect of individual modalities on first impressions}\label{sec:effect}

\begin{figure}[t!]
	\centering
	\includegraphics[width=0.48\textwidth]{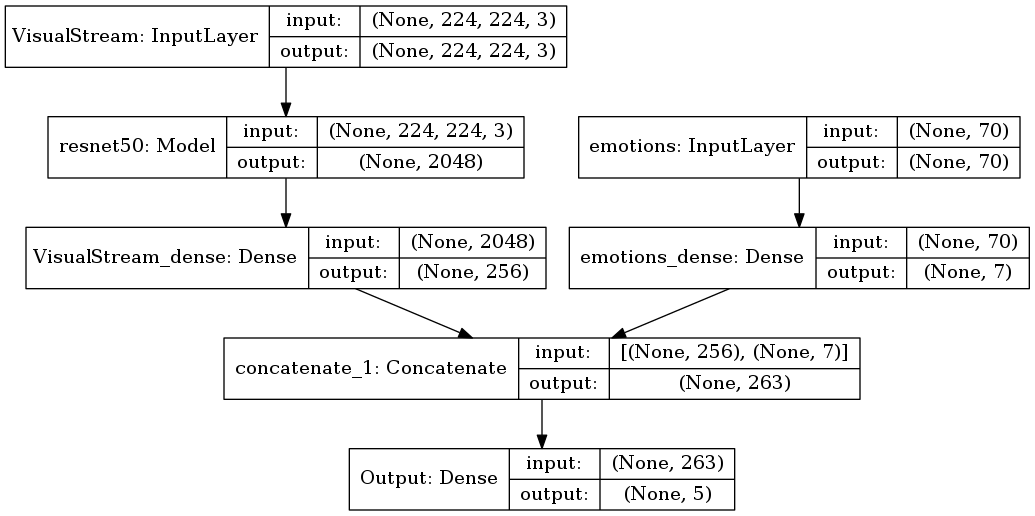}
\caption{Architecture of \textit{``Visual+Emotion''} experiment. Emotion is first encoded with a 70D histogram, concatenating information from first and second half of the video (Sec.~\ref{sub:emotion}) before late fusion.}
	\label{fig:video_emotions_arch}
\end{figure}

One of our goals is to assess the contribution of each information source, and how they can improve a baseline model that uses only raw visual information. 
To do so, we perform a series of experiments by training different versions of our model. As baseline, we use as input visual information only (i.e., raw images). Then, we incrementally add to this baseline model the additional factors and audio stream. In the experiments where audio is not considered, the dimensionality of the visual stream is set to 256. 
In particular, we test the following models: \textit{``Visual''} (i.e., our baseline model), \textit{``Visual+Factor''} (one for each factor) and \textit{``Visual+Audio''}. As an example, Figure \ref{fig:video_emotions_arch} illustrates the architecture of the \textit{``Visual+Emotion''} experiment. In that case, the visual cue is combined at a late stage with deep features associated to facial emotion expression before regressing the Big-Five traits.


\subsubsection{Dynamic factors analysis}
\label{sub:dynamic}

Before delving deeper into the contribution of each modality, we focus on analysing the influence that the choice of \textit{slice size/location} has on the performance of the factors that change over time, namely emotion and attractiveness. Particularly, we attempt to compare the performance of orderless versus ordered video statistics, and to assess whether a specific portion of a video can be more informative than others. 


For each dynamic factor we evaluate two aggregation methods: first, aggregating their individual predictions in a single video-level normalised histogram, that is, orderless consensus; and second, partitioning the video in two segments and aggregating the frame-level predictions of each segment separately with one histogram per segment (i.e., 2 concatenated histograms per video), thus obtaining an ordered consensus. Each experiment constitutes a different model combining baseline along with the specific factor using the evaluated temporal consensus approach.

As shown in Tables \ref{tab:temporal_emotion} and \ref{tab:temporal_attractiveness}, ordered consensus outperforms orderless, on average, for both emotion and attractiveness, being particularly beneficial to predict \textit{Openness to Experience}. This suggests that the order in which emotions are expressed (and perceived), as well as attractiveness is perceived, indeed matters when making personality judgments. In contrast, predicting \textit{Extraversion} from a single video-level representation is remarkably advantageous when considering the attractiveness descriptor, which could imply that it is an orderless trait.

{\renewcommand{\arraystretch}{1.5}
\begin{table}[t!]
\centering\scriptsize\setlength\tabcolsep{2pt}
\caption{Dynamic attributes: \textit{``Visual+Emotion''}. Different time intervals and representations for the temporal domain.}
\begin{tabular}{c|c|c|c|c|c|c|c}
\hline
\textbf{Input} & \textbf{Num. histograms} & \textbf{Avg.} & \textbf{O}  & \textbf{C} & \textbf{E}       & \textbf{A}     & \textbf{N} \\ \hline
Orderless & 1      & 0.9148   &	0.9144 &	0.9214	&   \textbf{0.9154}	&   \textbf{0.9132}&     0.9097  \\ \cline{1-8} 
                           Ordered  & 2      & \textbf{0.9152}   &	\textbf{0.9153} &	0.9214	&   0.9152	&   0.9131 &    \textbf{0.9110} \\ \hline \hline
First half & 1                                   & 0.9147   &	\textbf{0.9148} &	0.9211 &	\textbf{0.9155} &	0.9130 &	0.9092 \\ \hline
Second half& 1                                   & \textbf{0.9148}	&   0.9144 &	\textbf{0.9213} &	0.9149 &	\textbf{0.9133} &	\textbf{0.9100} \\ \hline
\end{tabular}
\label{tab:temporal_emotion}
\end{table}}

{\renewcommand{\arraystretch}{1.5}
\begin{table}[t!]
\centering\scriptsize\setlength\tabcolsep{1.5pt}
\caption{Dynamic attributes: \textit{``Visual+Attractiveness''}. Different time intervals and representations for the temporal domain.}
\begin{tabular}{c|c|c|c|c|c|c|c}
\hline
\textbf{Input} & \textbf{Num. histograms} & \textbf{Avg.}   & \textbf{O}      & \textbf{C}      & \textbf{E}      & \textbf{A}      & \textbf{N}      \\ \hline
Orderless    & 1  & 0.9147           &	0.9138          &	0.9208          &	\textbf{0.9158}   &	0.9130              &	0.9100  \\ \cline{1-8} 
                              Ordered  & 2  & \textbf{0.9148} &	\textbf{0.9147} &	\textbf{0.9210}           &	0.9151          &	\textbf{0.9133}              &	0.9100    \\ \hline \hline
First half                   & 1  & 0.9146           &	0.9136           &	\textbf{0.9212} &	0.9146          &	0.9133               &	0.9102 \\ \hline 
Second half                    & 1  & \textbf{0.9148} &	\textbf{0.9142}          &	0.9211          &	\textbf{0.9149}          &	\textbf{0.9136} &	\textbf{0.9104} \\ \hline
\end{tabular}
\label{tab:temporal_attractiveness}
\end{table}}

In addition, we analyse the use of either the first or the second half of the video to investigate their informative power on the temporal domain. For instance, if we take a look at the accuracy of the different time intervals (i.e., when using the first or the second half of the video) and in particular in the case of emotion expressions (Table~\ref{tab:temporal_emotion}), we can observe that both halves of the video convey, on average, similar information to code/decode apparent personality. For attractiveness (Table~\ref{tab:temporal_attractiveness}), however, the second half slightly outperforms the first half in terms of informative power for almost all traits, suggesting that we may need some time to consolidate our first impressions when taking attractiveness into account.

{\renewcommand{\arraystretch}{1.5}
\begin{table}[t!]
\centering\small\setlength\tabcolsep{2pt}
\caption{Dynamic modality: \textit{``Visual+Audio''}. Different time intervals and representations for the temporal domain.}
\begin{tabular}{c|c|c|c|c|c|c}
\hline
\textbf{Input}  & \textbf{Avg.}   & \textbf{O}      & \textbf{C}      & \textbf{E}      & \textbf{A}      & \textbf{N}      \\ \hline
Whole waveform & 0.9157 &	0.9157 &	0.9210 &	0.9156 &	0.9139 &	0.9122 \\ \hline \hline
First half (1A)  & \textbf{0.9160} &	\textbf{0.9162} &	0.9214 &    \textbf{0.9157} &	\textbf{0.9145} &	\textbf{0.9124} \\ \hline 
Second half (2A) & 0.9149 &	0.9146 &    \textbf{0.9219} &	0.9143 &	0.9124 &	0.9114 \\ \hline
\end{tabular}
\label{tab:temporal_audio}
\end{table}}

While not considered an attribute but rather an additional source for this work, raw audio inherently possesses temporally-ordered information, so it can also be treated as a dynamic factor. Consequently, we also evaluate different time intervals for the audio modality using the whole waveform in addition to each half. As one can observe in Table \ref{tab:temporal_audio}, it is clear that the first half alone contains all necessary information to better predict all traits, except for the case of \textit{Conscientiousness}, where the second half surpasses the rest of time intervals. Interestingly, using the whole video seems to perform an average among the informative power of both video fragments.







\subsubsection{Factors comparison}
\begin{figure*}[t!]
	\centering
	\includegraphics[width=\textwidth]{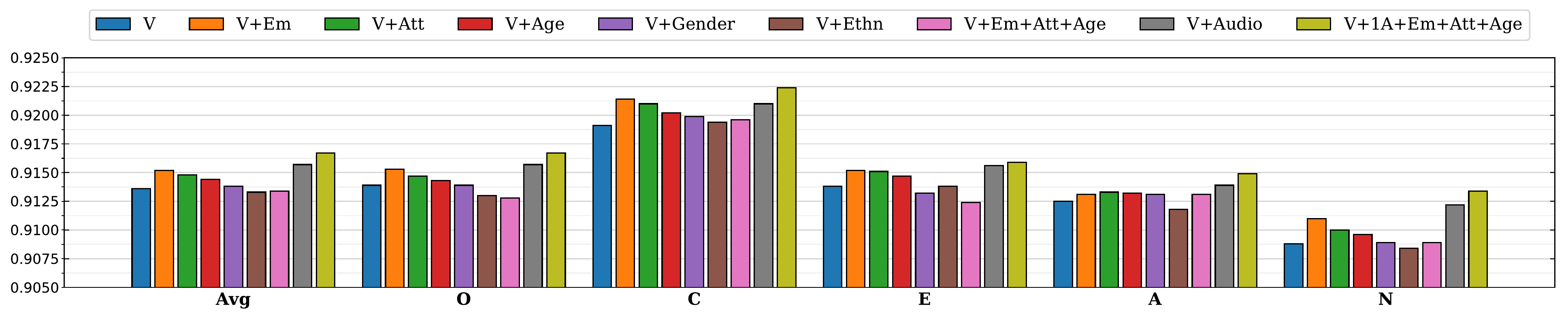}
	\vspace{-0.2cm}
	\caption{Comparison of our approaches combining different information modalities, grouped by Big-Five personality trait, in terms of mean accuracy.}
	\label{fig:trait_comparison}
\end{figure*}

The best performing models for emotion and attractiveness (i.e., ordered consensus) are selected for the comparison of the different combinations of modalities. Performance is shown in Figure \ref{fig:trait_comparison} for each model and apparent personality trait. 

In line with literature findings, adding emotion evidence in  \textit{``Visual+Emotion}'' (V+Em) substantially benefits the prediction of all traits, with \textit{Conscientiousness} and \textit{Neuroticism}\footnote{Normalised ground truth labels provided with the First Impression~\cite{ponce2016chalearn} database for \textit{Neuroticism} are in fact the inverse of Neuroticism, i.e., ``Emotion stability''. To keep the coherence with Big-Five model (i.e., OCEAN), such values were inverted again in our study and treated as Neuroticism.} showing the largest improvement. \textit{``Visual+Attractiveness''} (V+Att) comes in second, followed by \textit{``Visual+Age''} (V+Age). Both methods outperform the baseline, although to a lesser extent than emotions, only scoring slightly higher than the latter for \textit{Agreeableness}. \textit{``Visual+Gender''} (V+Gender), however, just shows a moderate improvement with respect to the baseline for all traits except \textit{Extraversion}, for which prediction ability is partly damaged. When adding ethnicity information to visual features in \textit{``Visual+Ethnicity''} (V+Ethn), we only observe small benefits for \textit{Extraversion} and \textit{Conscientiousness}, performing worse on average than when just using visual cues. This downward trend stops with the last tested modality, \textit{``Visual+Audio''} (V+Audio), being the best performing model among the individual modalities. Its effect is especially noticeable for \textit{Neuroticism}, which is the most difficult trait to predict on average (in line with~\cite{junior2018first}). With respect to personality traits, \textit{Conscientiousness} is the easiest predicted trait by all methods, followed at a distance by \textit{Openness to Experience} and \textit{Extraversion}, as demonstrated by previous state-of-the-art approaches (see Table \ref{tab:sotacomparison}).

\subsection{Multi-modal personality prediction and comparison with existing methods}
In this section, we evaluate the combination of all modalities in accordance with the experimental definition of Sec.~\ref{sec:effect}. Since adding gender and age partially harms performance, we discard them from our tested multi-modal approaches. Therefore, we first evaluate the joint combination of emotion, attractiveness and age with visual information, in \textit{``Visual+Emotion+Attractiveness+Age''} (V+Em+Att+Age). Second, we add the audio modality to further enrich the final representation. However, to boost performance, we just use the first half of the audio (represented by ``1A'' in Table~\ref{tab:temporal_audio}, and detailed in Sec.~\ref{sub:dynamic}), which has proved to be more informative for personality prediction. We refer to this last model as \textit{``Proposed''} (V+1A+Em+Att+Age). As illustrated in Figure \ref{fig:trait_comparison}, the \textit{``Proposed''} model surpasses all other approaches thanks to the combination of the ``best'' selected features, but in special to the inclusion of the audio modality. 
In this case, it can be observed the largest difference in accuracy for  \textit{Neuroticism} if compared to the V+Em+Att+Age model (i.e., a selection of the best features without audio). On the other hand, if compared to the baseline model (i.e., V), the V+Em+Att+Age model has positive effects on \textit{Conscientiousness}, \textit{Agreeableness} and \textit{Neuroticism}, but on average its prediction capability is lower than the baseline.

Next, we compare the \textit{``Proposed''} model to state-of-the-art alternatives on the FI dataset. Results are given in Table \ref{tab:sotacomparison}. We observe that our model outperforms most published results, and is on par with the state-of-the-art method of~\cite{kaya2017multi}. While on average their method achieves a slightly superior accuracy, our model is able to predict \textit{Conscientiousness} and \textit{Agreeableness} better than their method, being the latter one of the most difficult to predict. The approach of \cite{kaya2017multi} combines audio and visual features from both hand-crafted and deep features, and defines a fusion hierarchy adding kernel ELM and Random Forest classifiers on top, overall having a high complexity and reducing interpretability. On the contrary, our approach with task-specific models fosters explainability and is conceptually simple.

{\renewcommand{\arraystretch}{1.5}
\begin{table}[t!]
\centering
\caption{Comparison of performance on personality traits prediction with state-of-the-art methods using mean accuracy.}
\begin{tabular}{c|c|c|c|c|c|c}
\hline
\textbf{Method}                                   & \textbf{Avg.} & \textbf{O} & \textbf{C} & \textbf{E} & \textbf{A} & \textbf{N} \\ \hline
\textit{Proposed} &  0.9167 & 0.9167 & \textbf{0.9224} & 0.9159 & \textbf{0.9149} & 0.9134           \\ \hline
\cite{kaya2017multi}    & \textbf{0.9173}           & \textbf{0.9170}            & 0.9198                     & \textbf{0.9213}                & 0.9137                 & \textbf{0.9146}               \\ \hline
\cite{zhang2016deep}    & 0.9130           & 0.9124            & 0.9166                     & 0.9133                & 0.9126                 & 0.9100               \\ \hline
\cite{subramaniam2016bi} & 0.9121           & 0.9117            & 0.9119                     & 0.9150                & 0.9119                 & 0.9099               \\ \hline
\cite{guccluturk2016deep}    & 0.9109           & 0.9111            & 0.9138                     & 0.9107                & 0.9102                 & 0.9089               \\ \hline
\end{tabular}
\label{tab:sotacomparison}
\end{table}}

\subsection{Qualitative-quantitative analysis of contributions}

\begin{figure*}[t!]
	\centering
	\includegraphics[width=\textwidth]{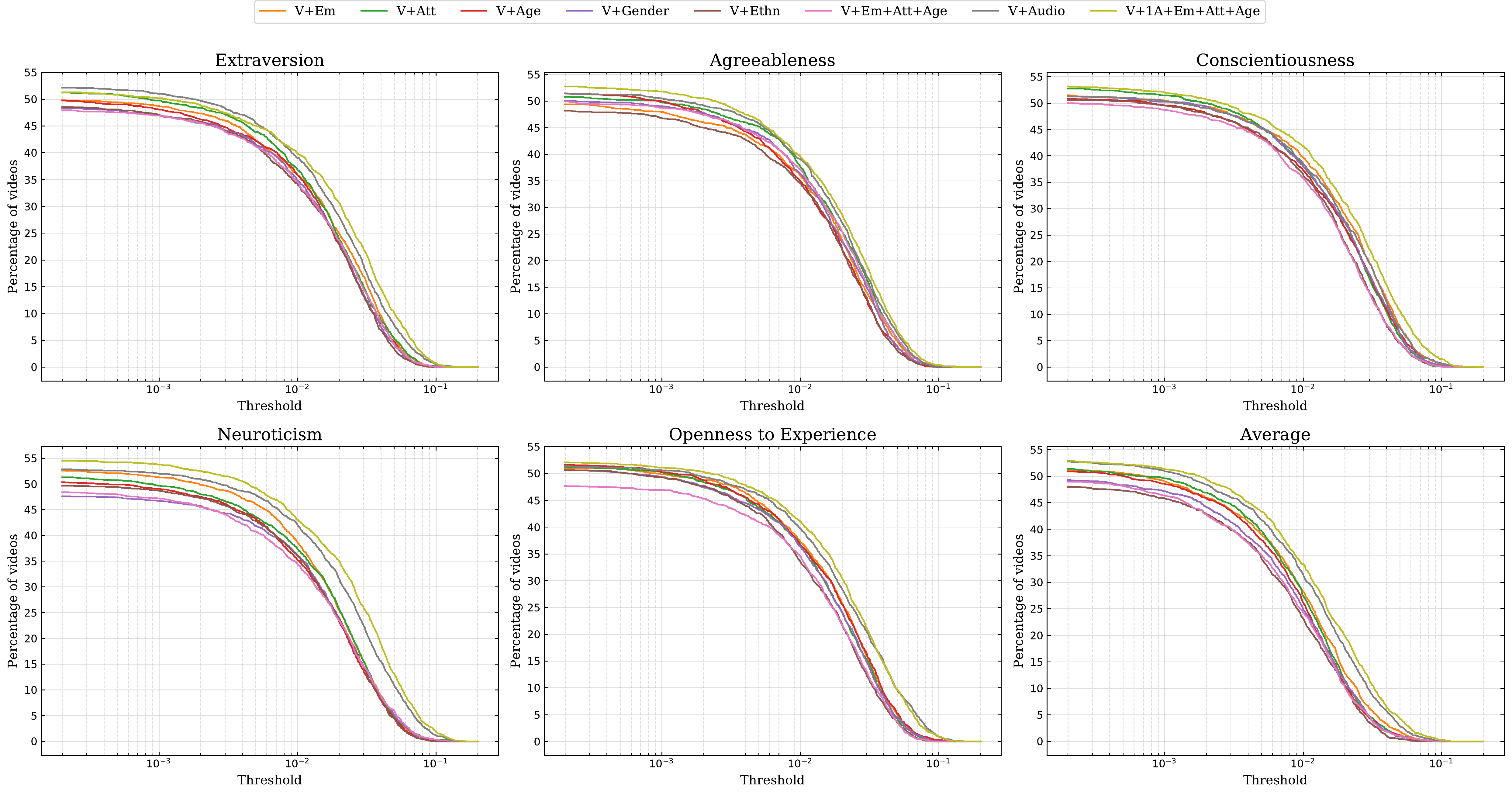}
	\caption{Percentage of videos whose accuracy improves when using a specific method compared to the baseline, as a function of the difference of prediction errors between baseline and selected method.}
	\label{fig:thresholds}
\end{figure*}

So far, we have demonstrated the improvement in performance of the proposed methods with respect to the baseline. To better understand how the additional features improve the final accuracy, we compare the differences between prediction and ground truth labels for both baseline and a selected method, and compute the residual of differences $d_{ij}$. Residuals greater than a threshold $th=0$ imply a better performance of the compared method against the baseline. To quantify the degree of improvement in terms of the percentage of videos for which a compared method improves the baseline, we use a moving threshold in Figure \ref{fig:thresholds} as:

\begin{equation}
\label{eq:diff}\scriptsize
	d_{ij} = \abs{p^{(b)}_{ij} - gt_{ij}} - \abs{p^{(c)}_{ij} - gt_{ij}}, 
    	\frac{\sum_{i=1}^{N} \mathds{1}\{d_{ij} >= th\}}{N}, \forall th \in [0,1],
\end{equation}
%

where $b$ corresponds to the baseline model (i.e., V), and $c$ to the specific compared method. Note that for higher threshold values on the residuals, a higher improvement is achieved by the \textit{``Proposed''} method compared to the rest of approaches, having the strongest effect on \textit{Conscientiousness} and \textit{Neuroticism}. 

\begin{figure*}[t!]
	\centering
	\includegraphics[width=\textwidth]{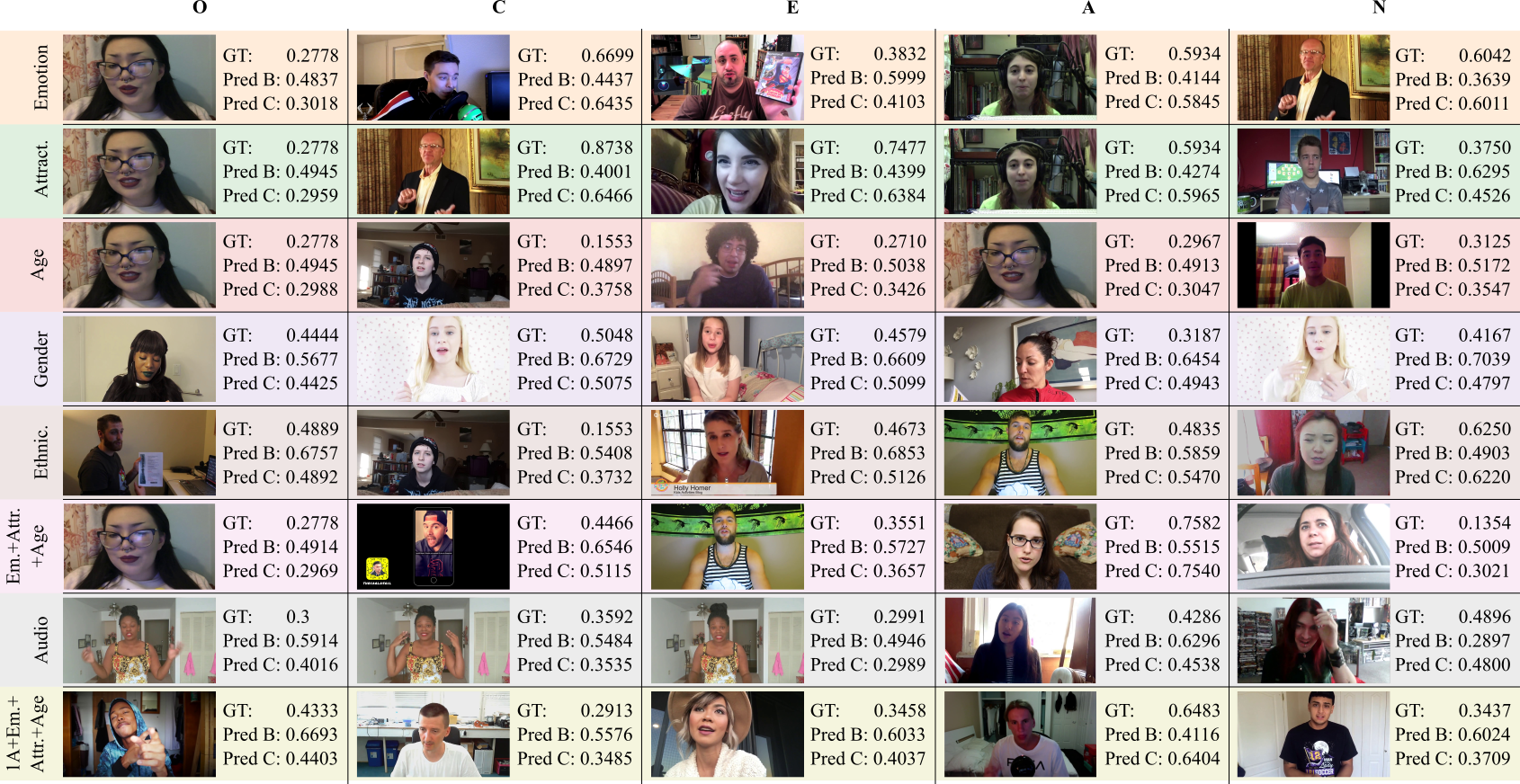}
	\vspace{-0.2cm}
	\caption{Examples of images from videos that maximise the improvement (per trait and method) obtained by a particular method when compared to the baseline (i.e., when combining visual information ``V'' with other modality). \textit{GT}: Ground truth label. \textit{Pred B}: Prediction from baseline. \textit{Pred C:} Prediction with compared method.}
	\label{fig:example_images}
\end{figure*}

Figure \ref{fig:example_images} shows examples of the videos that maximise this residual, for each method and trait. For instance, we notice that the addition of gender information helps predicting personality better for women, suggesting an implicit gender bias. We also observe that several videos appear multiple times, which implies that they are more difficult to predict just with raw visual features, and that the guidance provided to the network with the additional information is particularly useful for them. Finally, if we compare the ground truth labels against the predicted values for all modalities and personality traits, we can see an interesting effect. The baseline method tends to predict values concentrated in the centre of the distribution (i.e., around 0.5), whereas the compared methods are able to predict more extreme values. This indicates that the additional attributes help the model move away from a safer estimation.

\subsection{Analysing the influence of bias sources}\label{sec:attributes:analysis}
In this section, we analyse the correlation of attributes (i.e., gender, ethnicity, age, attractiveness and emotion expression) with respect to personality trait ground truth labels, in order to gain some insight into how these bias sources affect the annotations. Some of these attributes have already been studied~\cite{escalante2018explaining} on the FI~\cite{ponce2016chalearn} dataset (briefly discussed in Sec.~\ref{sec:bias}). Hence, our study complements previous analyses of bias factors on this data.

\subsubsection{Gender and Ethnicity}\label{subsec:gender:ethnicity}

\begin{figure}[t!]
	\centering
	\includegraphics[width=0.45\textwidth]{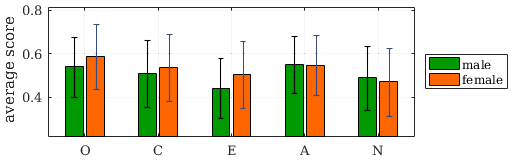}
	\vspace{-0.2cm}
	\caption{Average personality trait scores for males/females.}
	\label{fig:gender}
\end{figure}

In Figure~\ref{fig:gender}, we show the average score per trait/gender (for the FI database) obtained from the ground truth files. The database is somehow balanced with respect to gender (i.e., 45\% males and 55\% females), which benefits the analysis. As it can be seen, females obtained higher scores for almost all personality traits except for \textit{Agreeableness} (as reported in~\cite{escalante2018explaining}) and for \textit{Neuroticism}, which may indicate a difference in the target population of observed people or a gender bias from the annotators.

Table~\ref{tab:gender:ethnicity} shows the average trait obtained for different ethnicities and gender category, from where the following observations can be made: 1)~the database is extremely unbalanced with respect to ethnicity (i.e., 86\% of samples are composed of Caucasians); 2)~gender is unbalanced for Asians and Afro-Americans (i.e., around 70\% of samples for each of these categories are females); 3)~Asians and Caucasians follow the same trend as shown in Figure~\ref{fig:gender} (i.e., females are scored higher for most traits, except for \textit{Agreeableness} and \textit{Neuroticism}), however, a larger number of samples for Asian people should be used to draw a stronger discussion/analysis; 4)~differently from Asians and Caucasians, Afro-American males scored higher than females for \textit{Conscientiousness} and \textit{Agreeableness}, and lower for \textit{Neuroticism}, which indicates that a different target population might be sampled for such category or there may be some bias towards Afro-Americans; 5)~Afro-Americans were scored lower for almost all traits (and respective gender categories, expect for \textit{Neuroticism}) if compared to Asians and Caucasians, which may reflect some preconception towards this group.

{\renewcommand{\arraystretch}{1.5}
\begin{table}[t!]
\centering
\caption{Average personality traits on the First Impressions database. Distributions~(\%) per ethnicity and gender are shown.}
\setlength\tabcolsep{2.0pt}
\begin{tabular}{r|c|c|c|c|c|c}
 \hline
\textbf{Gender} & \textbf{\%} & \textbf{O} & \textbf{C} & \textbf{E} & \textbf{A} & \textbf{N} \\ \hline
\multicolumn{1}{l}{} & \multicolumn{6}{c}{\textbf{Asian} (3\%)}                                           \\ \hline
\textbf{Male}  & 29\% & 0.54$\pm$0.11 & 0.52$\pm$0.12 & 0.48$\pm$0.10 & 0.55$\pm$0.10 & 0.48$\pm$0.10 \\ \hline
\textbf{Female}  & 71\% & 0.59$\pm$0.14 & 0.54$\pm$0.15 & 0.52$\pm$0.14 & 0.55$\pm$0.13 & 0.47$\pm$0.15         \\ \hline  \hline
\multicolumn{1}{l}{} & \multicolumn{6}{c}{\textbf{Caucasian} (86\%)}                                           \\ \hline
\textbf{Male} & 48\% & 0.54$\pm$0.14 & 0.51$\pm$0.15 & 0.44$\pm$0.14 & 0.55$\pm$0.13 & 0.49$\pm$0.15 \\ \hline
\textbf{Female} & 52\%  & 0.60$\pm$0.15 & 0.54$\pm$0.14 & 0.51$\pm$0.15 & 0.55$\pm$0.14 & 0.46$\pm$0.16 \\ \hline  \hline
\multicolumn{1}{l}{} & \multicolumn{6}{c}{\textbf{Afro-American} (11\%)}                                           \\ \hline
\textbf{Male} & 31\% & 0.52$\pm$0.14 & 0.50$\pm$0.16 & 0.43$\pm$0.14 & 0.54$\pm$0.13 & 0.48$\pm$0.15 \\ \hline
\textbf{Female} & 69\%  & 0.52$\pm$0.13 & 0.48$\pm$0.14 & 0.45$\pm$0.14 & 0.51$\pm$0.12 & 0.51$\pm$0.14 \\ \hline
\end{tabular}
\label{tab:gender:ethnicity}
\end{table}
}

It must be emphasised that a further analysis with a larger and balanced database should be used for a stronger discussion on this topic.  Moreover, additional information such as gender, age, ethnicity and real personality, among others attributes, of both observers and observed people could be used to support a stronger analysis of personality perception and related biases~\cite{junior2018first}.

\subsubsection{Age}
\begin{figure}[t!]
	\centering
	\includegraphics[width=0.35\textwidth]{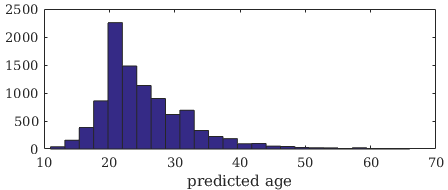}
	\vspace{-0.2cm}
	\caption{Distribution of predicted ages on the First Impressions database.}
	\label{fig:age}
\end{figure}

Next, we analyse the influence of age (of observed people) on personality perception. Although the analysis is based on predicted ages, as the FI dataset does not provide age labels, it gives a rough estimation about how such attribute affects personality perception. Figure~\ref{fig:age} shows the age distribution on the database. As one can observe, most values are centred on a particular age range, which may harm the analysis for those ages with small number of samples.

\begin{figure}[t!]
	\centering
	\includegraphics[width=0.45\textwidth]{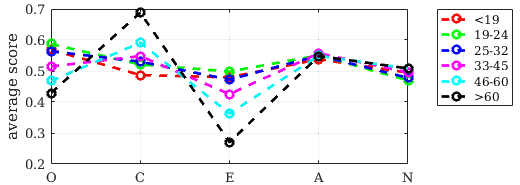}
	\vspace{-0.2cm}
	\caption{Average personality scores for different age-ranges.}
	\label{fig:age:plot}
\end{figure}

To analyse the influence of age, we first divided the database into 6 groups, based on different age-ranges (i.e., ``$<$19'',``19-24'',``25-32'',``33-45'',``46-60'' and ``$>$60''), as shown in Figure~\ref{fig:age:plot}. Then, we computed average personality scores for each group. From Figure~\ref{fig:age:plot}, some observations can be made: 1)~the perception of \textit{Openness} and \textit{Extraversion} decreases with respect to age, after a certain age range (i.e., when age $\geq$ ``33-45'' category), suggesting that older people are perceived to be less extroverted or open to new experiences; 2)~\textit{Conscientiousness} has a strong correlation with age, suggesting that the older one is, the more conscientious one is perceived; and finally, 3)~no meaningful trend was observed for other traits.

\subsubsection{Attractiveness}
\begin{figure*}[t!]
	\centering
	\includegraphics[width=\textwidth]{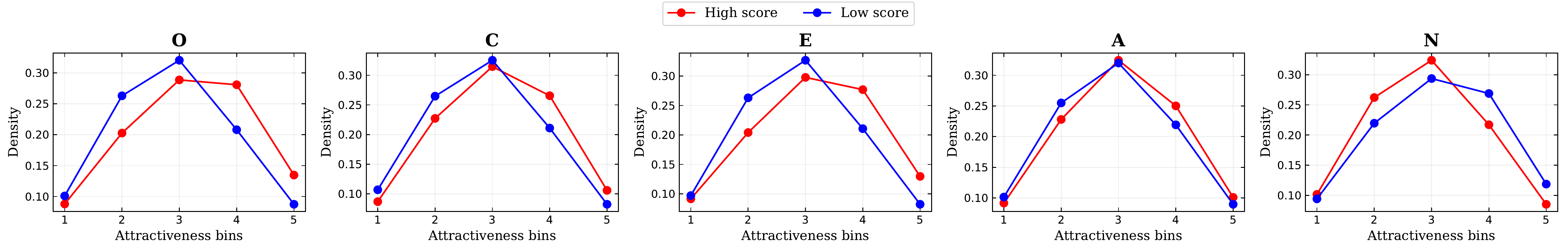}
	\caption{Correlation between attractiveness predictions (the higher the bin number, the more attractive one is perceived) and personality traits for the top 10\% individuals on the test set with higher (red curve) and lower (blue curve) scores on each trait.}
	\label{fig:attractive}
\end{figure*}

In this section, we analyse the attractiveness attribute. 
To do so, we select the 10\% of test samples with the highest and lowest ground truth labels for each personality trait, and correlate them to the obtained attractiveness predictions. Such values are represented as a normalised 5-bin histogram per video, in which the first bin represents a low attractiveness level, while the fifth represents a high level. Figure \ref{fig:attractive} shows that all histograms follow a Gaussian distribution, which was expected since the attractiveness model was trained with data following a similar Gaussian distribution from annotated attractiveness perception values. The analysis show that, overall, higher values of attractiveness perception correlate to higher values of all personality traits, except for \textit{Neuroticism}. In fact, few works in the field of personality computing have also reported a negative correlation between facial attractiveness and the perception of \textit{Neuroticism}~\cite{junior2018first,joshi2014automatic}.  
Moreover, our reported results are in line with the literature with respect to \textit{Conscientiousness}; being the most related Big-Five trait to \textit{intelligence}, it has been shown that perceptions of \textit{intelligence} are usually related to perceptions of attractiveness\cite{rohner2012recognition}, \cite{talamas2016eyelid}.

\subsubsection{Facial Emotion Expressions}\label{sec:emotion:bias}

The effect that face structure and emotion expressions have on personality perception has already been analysed in~\cite{guccluturk2017multimodal}. In their work, representative average images of those subjects that had the lowest and highest evaluations for each trait were created. According to their study, the observed differences were mostly related to face structure (e.g., femininity-masculinity of the faces) and facial expressions (e.g., a high score on \textit{Extraversion} trait and an associated smiling expression, which is aligned with the discussion presented in Section~\ref{sec:bias}). To extend the analysis of facial emotion expressions on personality perception (i.e., on the FI dataset), we followed the procedure described next.

\begin{figure*}[t!]
	\centering
	\subfigure{\includegraphics[height=2.4cm]{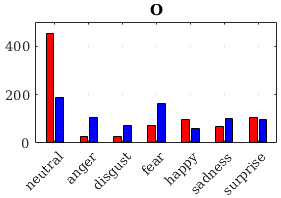}	}
	\subfigure{\includegraphics[height=2.4cm]{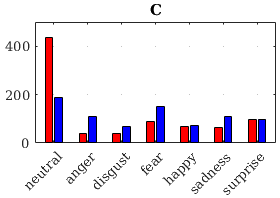}	}
	\subfigure{\includegraphics[height=2.4cm]{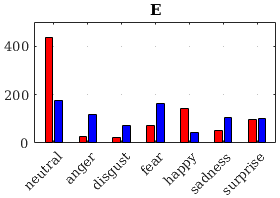}	}
	\subfigure{\includegraphics[height=2.4cm]{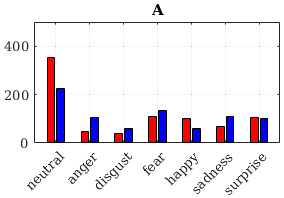}	}
	\subfigure{\includegraphics[height=2.4cm]{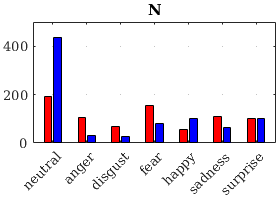}	}
    \caption{Accumulated frequency of emotions recognised with high probability (i.e., $\geq$70\% of accuracy) for the top 10\% individuals with higher (red bars) and lower (blue bars) scores for different personality traits.}
	\label{fig:personaliby:emotion}
\end{figure*}

First,  consider we have seven (one per emotion) 5-bin histograms per video, which represents the recognition accuracy (and frequencies) of estimated emotions over time (as described in Sec.~\ref{sub:emotion}).  Then, given an input video, we retrieve how frequently a particular emotion was recognised with accuracy higher than a predefined threshold (set to 70\% in our experiments, to deal with inaccuracies of the facial emotion expression recognition module). In Figure~\ref{fig:personaliby:emotion}, we show the accumulated frequencies per emotion obtained through such procedure, for the top 10\% individuals with higher (red bars) and lower (blue bars) scores on different personality traits, with respect to the whole database. This can give us an insight about what emotions were more frequently expressed by those individuals in a particular personality trait.

From Figure~\ref{fig:personaliby:emotion}, different observations can be made. First, \textit{anger}, \textit{disgust}, \textit{fear} and \textit{sadness} were in general more frequent for those individuals who scored low for almost all traits, except for \textit{Neuroticism}. In contrast, people with high values of \textit{Openness}, \textit{Extraversion} and \textit{Agreeableness} express the \textit{happy} emotion more frequently, which is even more salient for extroverted people. These results are consistent with evidences found in the literature~\cite{junior2018first}, suggesting that our facial expression recognition module was able to somehow model the personality-emotion relationship. No salient relationship pattern was observed between \textit{surprise} and personality traits. Finally, regarding the \textit{neutral} expression, it generally appeared with high frequency in all traits without any noticeable pattern, as it may be difficult for someone to express happiness or surprise, for instance, all the time in the video. However, people who scored high on different traits used to express the \textit{neutral} expression more frequently when compared to those individuals that scored low. Nevertheless, \textit{Neuroticism} showed exactly the opposite pattern, suggesting that people perceived low in \textit{Neuroticism} (i.e., having high \textit{Emotion stability}) tend to show more \textit{neutral} expressions than their counterparts.

\subsection{Limitations of the study}
The presented results are conditioned on the recognition performance of the emotion, gender, ethnicity, age and attractiveness attributes (Sec. \ref{sec:individual:models}), as well as on the quality of the dataset annotations (Sec. \ref{sec:dataset}). The labels are imperfect due to both the implicit biases of the personality perception task and the crowdsourcing process, in which few annotators just label a few videos. Further information of the annotators attributes would be required to better understand the biases present in data.

\section{Conclusion}\label{sec:conclusions}

We proposed a multi-modal deep learning approach for regressing the Big-Five apparent personality trait scores of people captured in one-person conversational video settings. We combined audio-visual data with automatically-recognised face attributes, including age, gender, ethnicity, facial expressions, and attractiveness, which allowed us to analyse person perception related bias in different ways, as well as their relationships with personality perception, being able to partially explain some of them. Furthermore, by combining such complementary information with audio-visual cues, also taking into account the temporal dimension of the data, we were able to achieve state-of-the-art results on the FI database. Given the large number of possible biases associated to first impressions, future work includes the analysis of additional factors, including upper-body gestures, clothing and scene understanding. Performing a multi-task learning would also benefit the joint learning of visual attributes and personality perception. Furthermore, although this work represents the most comprehensive analysis of possible sources of bias in personality perception from a computational perspective up to date, further research is required in order to include in these studies the effect of the observer attributes in relation to his/her perception biases. However, there is no publicly available dataset on this topic with rich and detailed information about both observers and observed people. Hence, the design and development of a new database covering this gap would definitively help to advance the state of the art on this research field.

\vspace{-0.3cm}

\ifCLASSOPTIONcompsoc
  \section*{Acknowledgments}
\else
  \section*{Acknowledgment}
\fi

This work has been supported by the Spanish projects TIN2015-66951-C2-2-R, TIN2016-74946-P (MINECO/FEDER,UE), CERCA Programme/Generalitat de Catalunya, and RTI2018-095232-B-C22 grant from the Spanish Ministry of Science, Innovation and Universities (FEDER funds). This work is partially supported by ICREA under the ICREA Academia programme. We gratefully acknowledge the support of NVIDIA Corporation with the donation of the GPU used for this research.
\vspace{-0.25cm}

\ifCLASSOPTIONcaptionsoff
  \newpage
\fi



\bibliographystyle{IEEEtran}

%
%


%
\vskip -2\baselineskip plus -1fil
\begin{IEEEbiography}[{\includegraphics[width=1in,height=1.25in,clip,keepaspectratio]{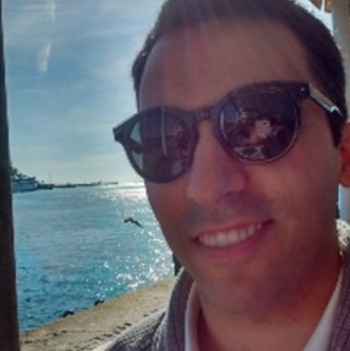}}]{Ricardo Dar\'io P\'erez Principi}
received his M.S. degree in Artificial Intelligence in 2019 from the Polytechnic University of Catalonia, Spain. He is currently working as a research collaborator within the Human Pose Recovery and Behavior Analysis (HUPBA) group at University of Barcelona (UB). His main research interests include multi-modal human behaviour analysis and machine learning.
\end{IEEEbiography}

\vskip -2\baselineskip plus -1fil
\begin{IEEEbiography}[{\includegraphics[width=1in,height=1.25in,clip,keepaspectratio]{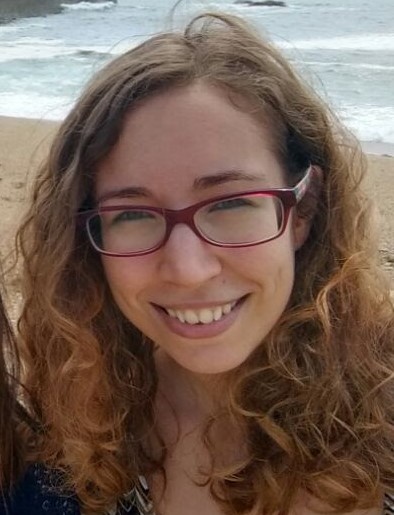}}]{Cristina Palmero} received her M.S. degree in Artificial Intelligence in 2014 from the Polytechnic University of Catalonia, Spain. She is currently a PhD student at Universitat de Barcelona (UB), and a member of the Human Pose Recovery and Behavior Analysis group (HuPBA), focusing on face-to-face group and dyadic interaction and automatic gaze estimation with remote cameras. Her main research interests include multi-modal human behaviour analysis and social signal processing.
\end{IEEEbiography}

\vskip -2\baselineskip plus -1fil
\begin{IEEEbiography}[{\includegraphics[width=1in,height=1.25in,clip,keepaspectratio]{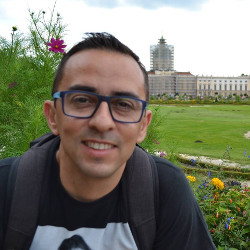}}]{Julio C. S. Jacques Junior} is a postdoctoral researcher at the Computer Science, Multimedia and Telecommunications department at Universitat Oberta de Catalunya (UOC), within the Scene Understanding and Artificial Intelligence (SUNAI) group. He also collaborates within the Computer Vision Center (CVC) and Human Pose Recovery and Behavior Analysis (HUPBA) group at Universitat Autonoma de Barcelona (UAB) and University of Barcelona (UB), as well as within ChaLearn Looking at People.
\end{IEEEbiography}

\vskip -2\baselineskip plus -1fil
\begin{IEEEbiography}[{\includegraphics[width=1in,height=1.25in,clip,keepaspectratio]{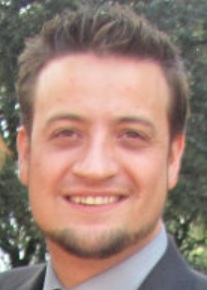}}]{Sergio Escalera} leads the Human Pose Recovery and Behavior Analysis (HuPBA) group at University of Barcelona (UB) and Computer Vision Center (CVC). He is an associate professor at UB, and member of CVC. He is vice-president of ChaLearn and chair of IAPR TC-12: Multimedia and visual information systems. He has been awarded with ICREA Academia. His research interests include affective and personality computing, with special interest in human pose recovery and behaviour analysis from multi-modal data.
 \end{IEEEbiography}







\end{document}